\documentclass{article} 
\usepackage{iclr2020_conference,times}

\usepackage{amsmath,amsfonts,bm}

\def\eqref#1{equation~\ref{#1}}

\def\1{\bm{1}}

\DeclareMathAlphabet{\mathsfit}{\encodingdefault}{\sfdefault}{m}{sl}
\SetMathAlphabet{\mathsfit}{bold}{\encodingdefault}{\sfdefault}{bx}{n}

\usepackage{hyperref}
\usepackage{cleveref}
\usepackage{url}
\usepackage{graphicx}
\usepackage[final]{changes} 
\definechangesauthor[name={Sim.~S.},color={blue}]{sims}
\definechangesauthor[name={D.~D.}, color={red}]{dd}
\definechangesauthor[name={B.~L.}, color={green}]{bl}
\def\sys/{TrueBranch}

\title{\sys/: Metric Learning-based Verification of Forest Conservation Projects}
\author{Simona Santamaria$^*$, David Dao$^*$, Ce Zhang  \\
Department of Computer Science\\
ETH Zurich\\
Zurich, Switzerland \\
\texttt{\{ssimona,david.dao,ce.zhang\}@inf.ethz.ch} \\
\And
Bj\"orn L\"utjens\thanks{Authors have contributed equally. } \\
Department of Aeronautics and Astronautic \\
Massachusetts Instit\textbf{}ute of Technology \\
Cambridge, USA \\
\texttt{lutjens@mit.edu} \\
}

\iclrfinalcopy 
\begin{document}

\maketitle

\begin{abstract}
International stakeholders increasingly invest in offsetting carbon emissions, for example, via issuing Payments for Ecosystem Services (PES) to forest conservation projects. Issuing trusted payments requires a transparent monitoring, reporting, and verification (MRV) process of the ecosystem services (e.g., carbon stored in forests).
The current MRV process, however, is either too expensive (on-ground inspection of forest) or inaccurate (satellite). Recent works propose low-cost and accurate MRV via automatically determining forest carbon from drone imagery, collected by the landowners. The automation of MRV, however, opens up the possibility that landowners report untruthful drone imagery. To be robust against untruthful reporting, we propose \sys/, a metric learning-based algorithm that verifies the truthfulness of drone imagery from forest conservation projects.
\sys/ aims to detect untruthfully reported drone imagery by matching it with public satellite imagery. Preliminary results suggest that nominal distance metrics are not sufficient to reliably detect untruthfully reported imagery. \sys/ leverages metric learning to create a feature embedding in which truthfully and untruthfully collected imagery is easily distinguishable by distance thresholding.
\vspace{-.16in}
\end{abstract}

\section{Introduction}
\label{introduction}

Agriculture, forestry, and other land use is a key driver of climate change, accounting for 23\% ($12.0 \pm 2.9$ GtCO2eq $\text{yr}^{-1}$) of total anthropogenic emissions of greenhouse gases during 2007-2016~\citep{IPCC_land_use_2019}, largely driven by deforestation and forest degradation. Deforestation does not only release carbon (e.g., through  slash-and-burn), but also destroys a multitude of other forest ecosystem services: preserving biodiversity, counteracting flooding and soil erosion, filtering water, and offering a livelihood for the local population. 

The causes of deforestation are mostly economically driven: expansion of commercial or subsistence agriculture, logging, fuelwood collection, or livestock grazing~\citep{Hosonuma_2012}). To counteract the economic incentives, payments for ecosystem services (PES)~\citep{Wunder_2007} are increasingly~\citep{Donofrio_2019} provided to forest conserving or restoring landowners 
by international stakeholders (e.g., through the governmental UN-REDD program~\citep{Gibbs_2007} or the commercial voluntary carbon market~\citep{Donofrio_2019}). 
However, current methods for monitoring, reporting, and verification (MRV) of the landowner-provided forest ecosystem services are either based on 1) on-ground inspection, which is too expensive (USD 20-30k), delayed (up to two years), corruptible, and biased~\citep{GoldStandard_2017}, 2) satellite, which is low-cost, but limited to the binary verification of forest/no-forest cover~\citep{Hansen_2013}, or 3) drones.

Recent works haved proposed low-cost and accurate MRV via drones~\citep{Dao_2019, daogainforest, Lutjens_2019}. Specifically, these works propose algorithms that estimate forest ecosystem services (e.g., stored forest carbon) from drone imagery that was reported by landowners.
Replacing on-ground inspection with remote assessment via drones, however, opens up the possibility of untruthfully reported imagery. Given that the landowner is financially incentivized by PES to report higher forest ecosystem services value (e.g., higher forest cover or biodiversity), 
the possibility of false reporting is high, but not addressed by previous works.
We hypothesize that the landowner could report untruthful imagery by altering the image 1) in  location, 2) in time, or 3) with adversarial perturbation (e.g., via PGD \citep{Madry_2018}), as displayed in~\Cref{fig:attack}.

\begin{figure}
  \centering
  \includegraphics[width=0.75\textwidth]{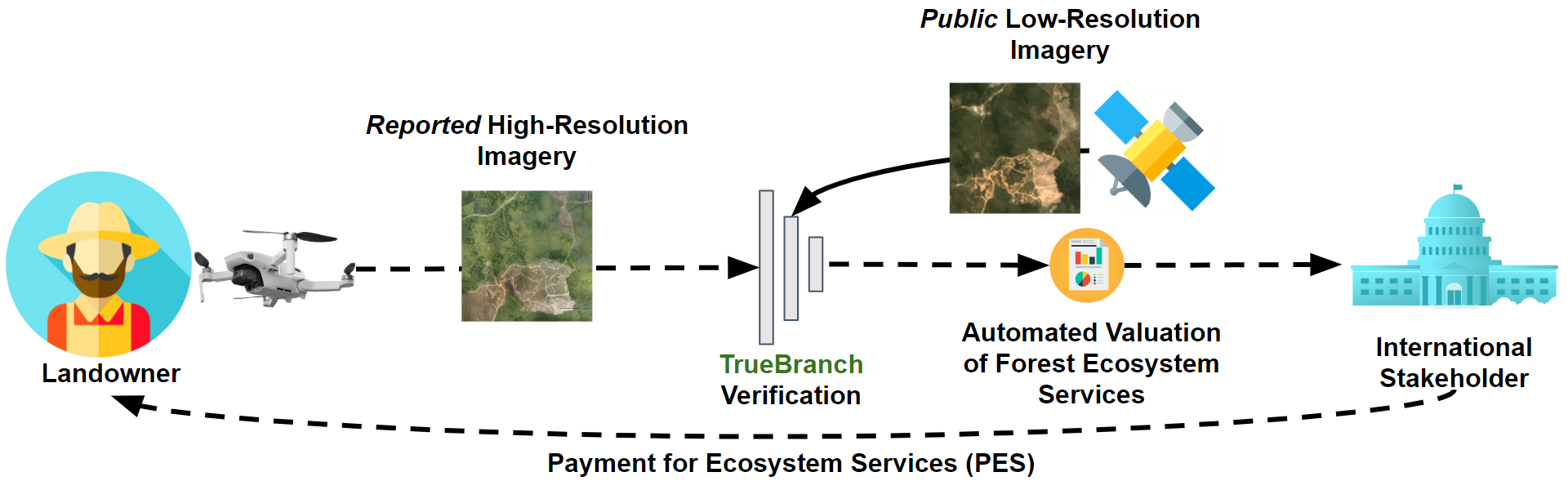}
  \vspace{-.15in}
  \caption{An overview of \sys/. A landowner takes a high-resolution drone image of their forest and reports the image, time stamp, and location. A metric learning-based algorithm, \sys/, verifies the submission with the corresponding low-resolution public satellite image.
  Another algorithm estimates the forest ecosystem service value based on the verified imagery and international stakeholders provide payments for ecosystem services (PES) to the landowner.} 
  \vspace{-.15in}
  \label{fig:overview}
\end{figure}

We propose \sys/, a metric learning-based algorithm that verifies the trustworthiness of reported drone imagery of forest conservation projects. Specifically, \sys/ aims to verify the truthfulness of drone images via matching them with public satellite images. Matching is proposed to be done in a deeply learned feature space 
that, ideally, 1) allows for easy distinction of images with different value of forest ecosystem services, 2) is robust to adversarial perturbation of the drone image, and 3) generalizes the verification of drone images to other ecosystems (e.g., mangroves or peatlands). 
~\Cref{fig:overview} shows how \sys/ is embedded in the scheme of automated MRV systems to achieve low-cost, accurate, and trustworthy MRV, which will promote international investments in forest conservation. 
\vspace{-.1in}

\label{rel_work}
\section{Attack vectors}
\vspace{-.1in}
\label{attack vectors}

\begin{figure}
  \centering
  \includegraphics[width=0.75\textwidth]{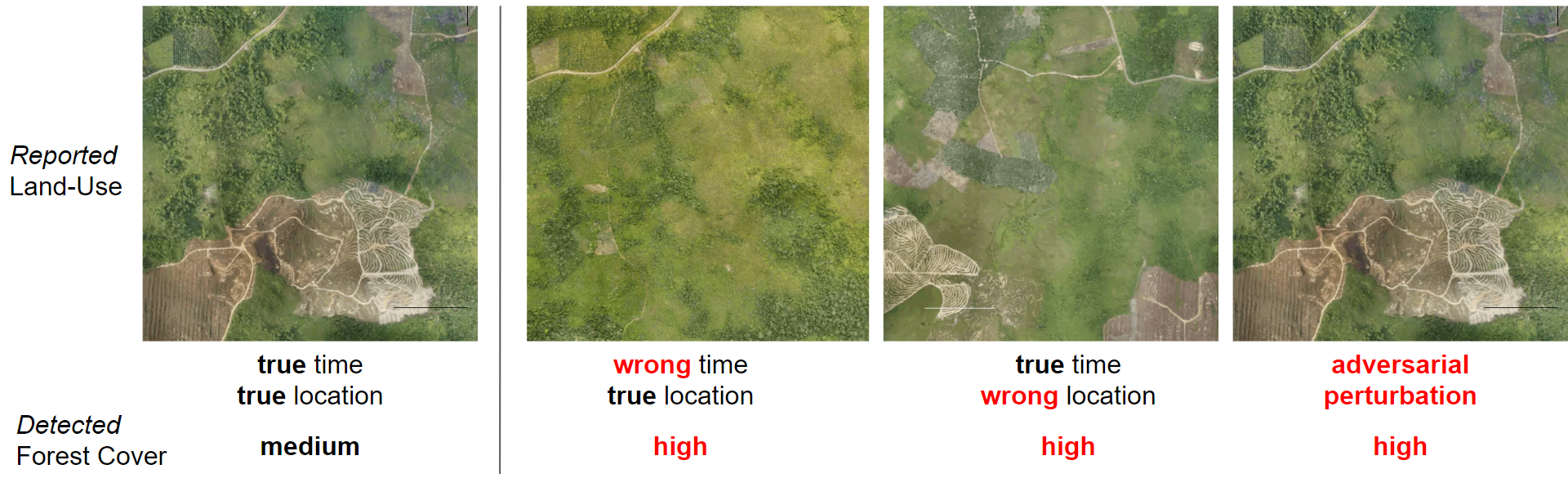}
  \vspace{-.15in}
  \caption{An overview of possible attack vectors in time, location, and value (left-to-right) that trick an automated forest valuation algorithm to detect high forest cover. 
}
  \label{fig:attack}
  \vspace{-.2in}
\end{figure}
Landowners are financially incentivized to report untruthful drone imagery that displays forest with higher ecosystem services value to receive higher PES. To reason about an algorithm that detects untruthful imagery, we classify common attack vectors with examples, as displayed in~\Cref{fig:attack}:
\begin{enumerate}
    \item Altering drone image \textbf{location:} The landowner has land with 50\% forest cover and reports imagery from a neighbouring land with 80\% forest cover to receive higher valuation.
    \vspace{-.06in}
    \item Altering drone image \textbf{time stamp:} The landowner reports imagery (e.g., with altered time stamp metadata) from previous flights, before their land has been logged or cleared.
        \vspace{-.06in}
\item Altering drone image \textbf{values}: The landowner tricks a neural network-based forest valuation algorithm into estimating higher ecosystem value by altering the image values with sophisticated attacks such as human-imperceptible \textbf{adversarial perturbation} (e.g via PGD \citep{Madry_2018}). 
    \vspace{-.06in}
    \item A combination of the above.
        \vspace{-.06in}

\end{enumerate}

\section{Approach}

\subsection{Challenges of detecting attack vectors with nominal metrics}\label{sec:3_1}
\begin{figure}
  \centering
  \includegraphics[width=0.95\textwidth]{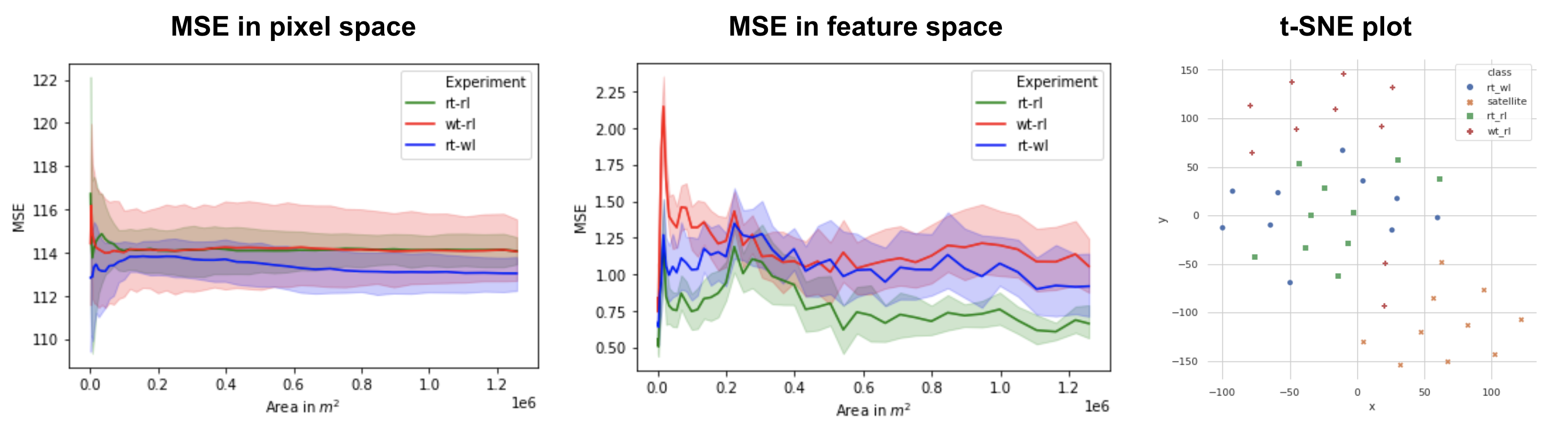}
  \vspace{-.2in}
  \caption{
  Preliminary results show that neither the MSE metric in pixel space nor in RESISC-45~\citep{neumann2019indomain} feature space are sufficient to reliably distinguish right time/right location (rt-rl) images from their attack vectors (wrong time/right location (wt-rl) and right time/wrong location (rt-wl)). The t-SNE plot shows that the drone images are naturally clustered together and difficult to separate. The plots have been generated using the 10 images from~\Cref{fig:images}.}
  \vspace{-.15in}
  \label{fig:prelim_results}
\end{figure}

We ran preliminary experiments to investigate if we can detect attack vectors (i.e., untruthfully reported imagery).~\Cref{fig:images} shows all used publicly available drone (0.3m/px) and satellite (4m/px) imagery from OpenForest and PlanetLabs, respectively.~\Cref{fig:prelim_results} suggests that nominal distance metrics (here, MSE in pixel or RESICS-45 feature space (i.e., the activation layer of a classification network, trained on satellite imagery~\citep{neumann2019indomain})) 
are not sufficient to reliably separate truthful from untruthful drone imagery by using the corresponding low-resolution satellite imagery.

\label{prelim_results}

\begin{figure}
  \centering
  \includegraphics[width=0.8\textwidth]{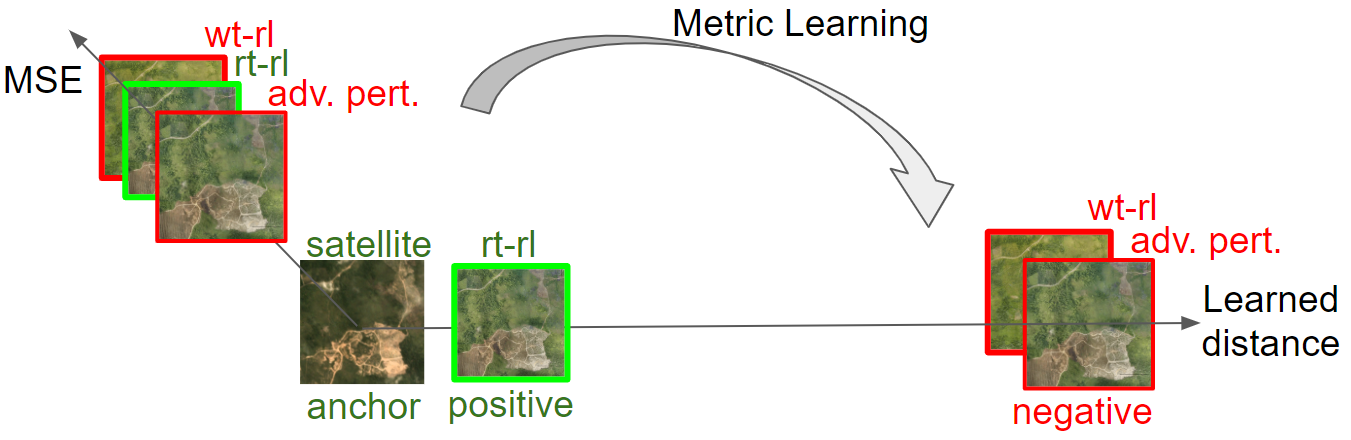}
  \vspace{-.15in}
  \caption{Illustration of metric learning. Truthful (green), untruthful (red), and satellite imagery are all close to each other, as measured by MSE distance. Metric learning pulls the satellite (anchor) and corresponding drone imagery (positive) even closer together and pushes the attack vectors (negative) away from the satellite imagery.
  }
  \vspace{-.2in}
  \label{fig:metric}
\end{figure}

\subsection{Detecting attack vectors with a learned metric}
Based on our preliminary results, we design \sys/ to leverage the triplet loss function~\citep{Schroff_2015} from the field of metric learning. The loss function is designed to pull truthful images closer to the true satellite image and push untruthful images away from it. The result of training with triplet loss should be a deeply learned feature space (i.e., learned metric) in which truthful and untruthful imagery is separated by a large margin, as displayed in~\Cref{fig:metric}.
In comparison to the nominal metrics from~\cref{sec:3_1}, the learned metric is also designed to be robust to adversarially perturbed images (sec. 2.3), as shown by~\cite{mao2019metric}. 
As metric learning requires large amounts of imagery and high-resolution drone imagery is scarce, future works could augment drone imagery by satellite imagery, enhanced by super resolution algorithms~\citep{deudon2020highresnet}.
Note that the metric learning-approach enables \sys/ to be extended from the verification of imagery of forest conservation projects to a multitude of other ecosystem conservation projects (e.g., mangroves, peatlands, wetlands, etc.) by adding the respective imagery.

\section{Conclusion}
\label{conclusion}
Truthfully reported imagery is a vital requirement to issue Payments for Ecosystem Services (PES) based on trusted and  automatic valuation of ecosystem services (e.g., stored forest carbon). 
In this paper, we list possible attack vectors and propose \sys/, a metric learning-based algorithm that uses public satellite data and metric learning to distinguish truthfully reported from untruthfully reported imagery.

\clearpage
\subsubsection*{Acknowledgments}
The authors are thankful for the guidance and advise by the mentors (Prof. Dava Newman, Forrest Meyen, Adam de Sola Pool), support from the local community (Sandro Pimentel, La Niebla Forest), academic collaborators (Prof. Pedro Brancalion, Prof. Paulo Guilherme Molin), non-governmental institutions (WWF Peru and Brazil) and CONAF (Daniel Montaner, Cesar Mattar, Jose Antonio Prado). Part of this research has been developed as part of the OpenSurface platform and a real-world pilot in Chile, which was launched at the COP25 United Nation’s Climate Summit. OpenSurface is funded by IDBLab and EIT Climate-KIC.

\bibliography{main}
\bibliographystyle{iclr2020_conference}

\clearpage
\appendix
\section{Appendix}
\begin{figure}[ht]
  \centering
  \includegraphics[width=0.50\textwidth]{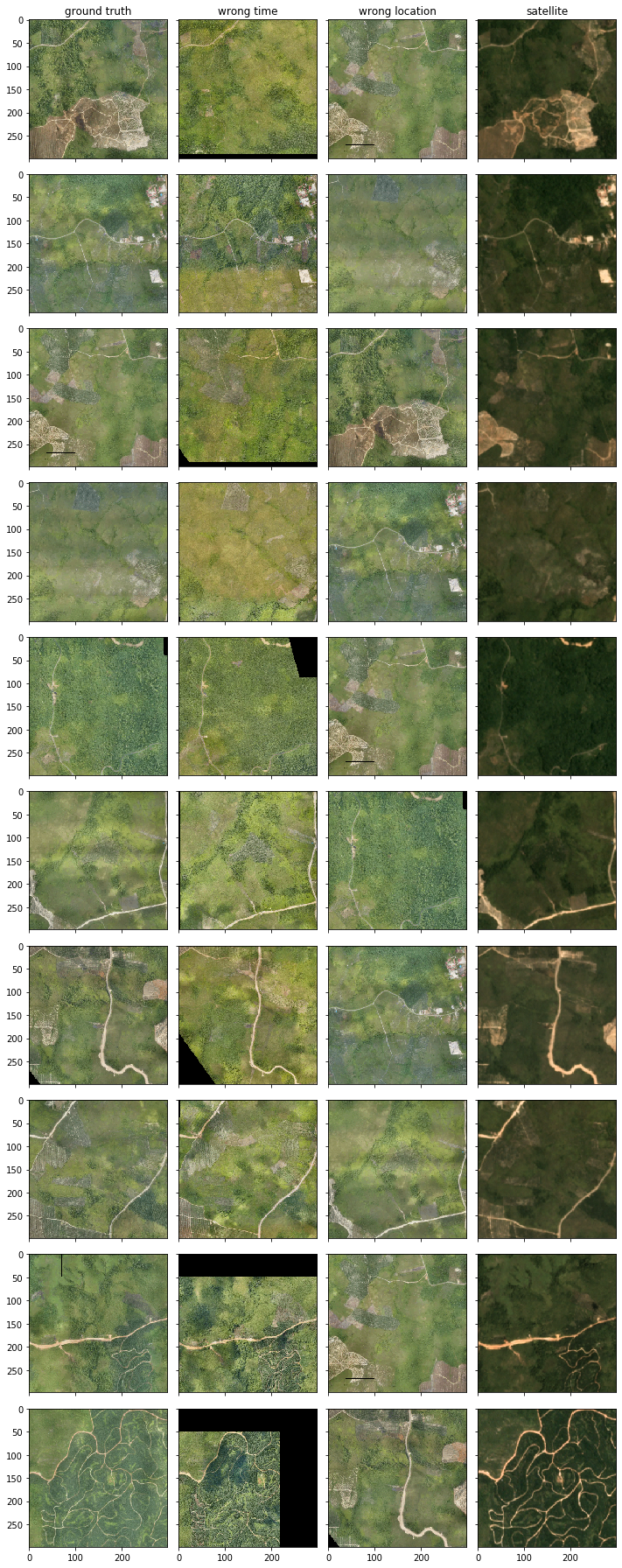}
  \caption{Drone and satellite images. The columns show drone imagery (from left to right), that is recorded 1) at the right time/right location, 2) wrong time/right location, 3) right time/wrong location, and 4) the ground truth satellite imagery. The x- and y-axis are in pixels. It is noteworthy that wrong time imagery only occasionally has higher forest cover (e.g., 1st or 3rd row), shows artifacts (e.g., last row), and generally has higher intensity than ground truth imagery. The wrong location images are the same as ground truth images, but in a different order. The satellite images have generally lower intensity and resolution than the ground truth imagery.}
  \label{fig:images}
\end{figure}
\end{document}